\documentclass[10pt,twocolumn,letterpaper]{article}

\usepackage{cvpr}
\usepackage{times}
\usepackage{epsfig}
\usepackage{graphicx}
\usepackage{amsmath}
\usepackage{amssymb}
\usepackage{listings}
\usepackage{makecell}
\usepackage{booktabs}
\usepackage{multirow}
\usepackage{color}
\usepackage{comment}
\usepackage{icomma}
\usepackage{enumitem}
\usepackage{amsthm}
\usepackage{authblk}
\setitemize{noitemsep,topsep=0pt,parsep=0pt,partopsep=0pt}
\setenumerate{noitemsep,topsep=0pt,parsep=0pt,partopsep=0pt}
\usepackage[bf,font=footnotesize]{caption}
\usepackage{subfig}
\author[2,3]{Junjie Zhang\thanks{The work was done while visiting The University of Adelaide.}}
\author[1]{Qi Wu\thanks{The first two authors contributed to this work equally.}}
\author[1]{Chunhua Shen}
\author[2]{Jian Zhang}
\author[3]{Jianfeng Lu}
\author[1]{Anton van den Hengel}
\affil[1]{Australian Centre for Robotic Vision, The University of Adelaide, Australia}
\affil[2]{Faculty of Engineering and Information Technology, University of Technology Sydney, Australia}
\affil[3]{School of Computer Science and Technology, Nanjing University of Science and Technology, China}

\usepackage{algorithm}
\usepackage{algorithmicx}
\usepackage{algcompatible}
\usepackage[noend]{algpseudocode}

\usepackage[pagebackref=true,breaklinks=true,letterpaper=true,colorlinks,bookmarks=false]{hyperref}

 \cvprfinalcopy 

\def\cvprPaperID{187} 

\ifcvprfinal\fi
\begin{document}

\title{Asking the Difficult Questions: Goal-Oriented Visual Question Generation \\via Intermediate Rewards}

\maketitle

\begin{abstract}


Despite significant progress in a variety of vision-and-language problems, developing a method capable of asking intelligent, goal-oriented questions about images is proven to be an inscrutable challenge. Towards this end, we propose a Deep Reinforcement Learning framework based on three new intermediate rewards, namely \textbf{goal-achieved}, \textbf{progressive} and \textbf{informativeness} that encourage the generation of succinct questions, which in turn uncover valuable information towards the overall goal. By directly optimizing for questions that work quickly towards fulfilling the overall goal, we avoid the tendency of existing methods to generate long series of insane queries that add little value. We evaluate our model on the GuessWhat?! dataset and show that the resulting questions can help a standard Guesser identify a specific object in an image at a much higher success rate.
\end{abstract}

\vspace{-20pt}
\section{Introduction}

\textit{Judge a man by his questions rather than by his answers.}
\begin{flushright}\textit{-Voltaire}\end{flushright}


Although Visual Question Answering (VQA)~\cite{antol2015vqa,Wu_2016_CVPR,xu2016ask} has attracted more attention, Visual Question Generation (VQG) is a much more difficult task. Obviously, generating facile, repetitive questions represents no challenge at all, but generating a series of questions that draw out useful information towards an overarching goal, however, demands consideration of the image content, the goal, and the conversation thus far. It could, generally, also be seen as requiring consideration of the abilities and motivation of the other participant in the conversation.

A well-posed question extracts the most informative answer towards achieving a particular goal, and thus reflects the knowledge of the asker, and their estimate of the capabilities of the answerer. Although the information would be beneficial in identifying a particular object in an image, there is little value in an agent asking a human about the exact values of particular pixels, the statistics of their gradients, or the aspect ratio of the corresponding bounding box. The fact that the answerer is incapable of providing the requested information makes such questions pointless. Selecting a question that has a significant probability of generating an answer that helps achieve a particular goal is a complex problem.

\begin{figure}[t]
	\centering
	\includegraphics[width=1\linewidth]{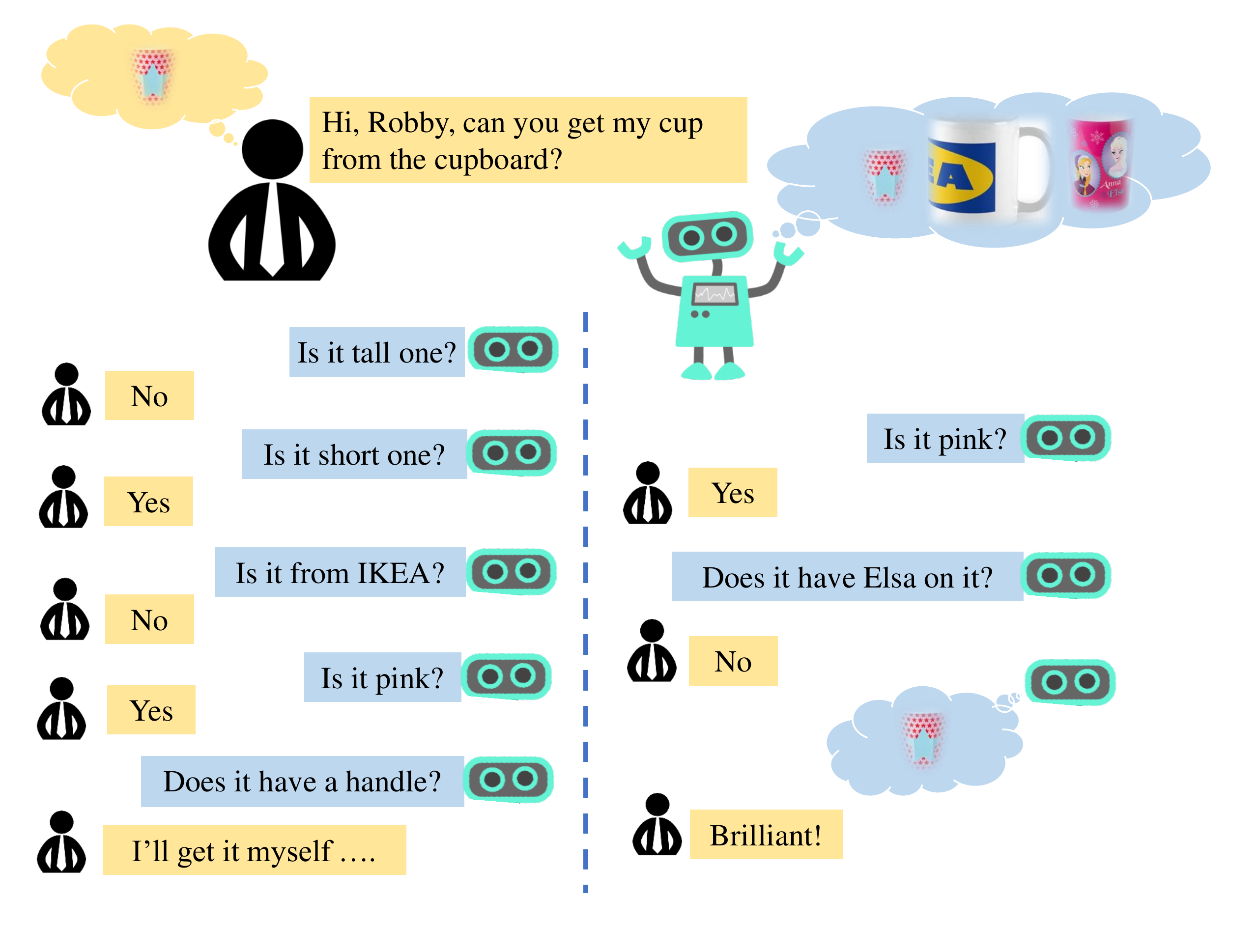}
	\vspace{-20pt}
	\caption{Two illustrative examples of potential conversations between a human and a robot. The left conversation clearly makes people frustrated while the right one makes people happy because the robot achieves the goal in a quicker way via less but informative questions.
	}
	\vspace{-18pt}
	\label{qa_demo}
\end{figure}

Asking questions is an essential part of the way humans communicate and learn. Any intelligent agent that seeks to interact flexibly and effectively with humans thus needs to be able to ask questions. The ability to ask intelligent questions is even more important than receiving intelligent, actionable answers. A robot, for example in Fig. \ref{qa_demo}, has been given a task and realized that it is missing critical information required to carry it out, needs to ask a question. It will have a limited number of attempts before the human gets frustrated and carries out the task themselves. This scenario applies equally to any intelligent agent that seeks to interact with humans, as we have surprisingly little tolerance for agents that are unable to learn by asking questions, and for those that ask too many.

As a result of the above, Visual Question Generation (VQG) has started to receive research attention, but primarily as a vision-to-language problem~\cite{li2017visual,mora2016towards, zhang2016automatic}. Methods that approach the problem in this manner tend to generate arbitrary sequences of questions that are somewhat related to the image~\cite{mostafazadeh2016generating}, but which bare no relationship to the goal. This reflects the fact that these methods have no means of measuring whether the answers generated to assist in making progress towards the goal. Instead, in this paper, we ground the VQG problem as a goal-oriented version of the game - GuessWhat?!, introduced in~\cite{guesswhat_game}. The method presented in~\cite{guesswhat_game} to play the GuessWhat game is made up of three components: the \texttt{Questioner} asks questions to the \texttt{Oracle}, and the \texttt{Guesser} tries to identify the object that the {Oracle} is referring to, based on its answers. The quality of the generated questions is thus directly related to the success rate of the final task.

Goal-oriented training that uses a game setting has been used in visual dialog generation previously~\cite{das2016visual,das2017learning}. However, these work focus on generating more human-like dialogs, not on helping the agent achieve the goal through better question generation. Moreover, previous work \cite{end_to_end_gw} only uses the final goal as the reward to train the dialog generator, which might be suitable for dialog generation but is a rather weak and undirected signal by which to control the quality, effectiveness, and informativeness of the generated question in a goal-oriented task. In other words, in some cases, we want to talk to a robot because we want it to finish a specific task but not to hold the meaningless boring chat. Therefore, in this paper, we use intermediate rewards to encourage the agent to ask short but informative questions to achieve the goal. Moreover, in contrast to previous works that only consider the overall goal as the reward, we assign different intermediate rewards for each posed question to control the quality. 

\begin{figure*}[t]
	\centering
	\includegraphics[width=1\linewidth]{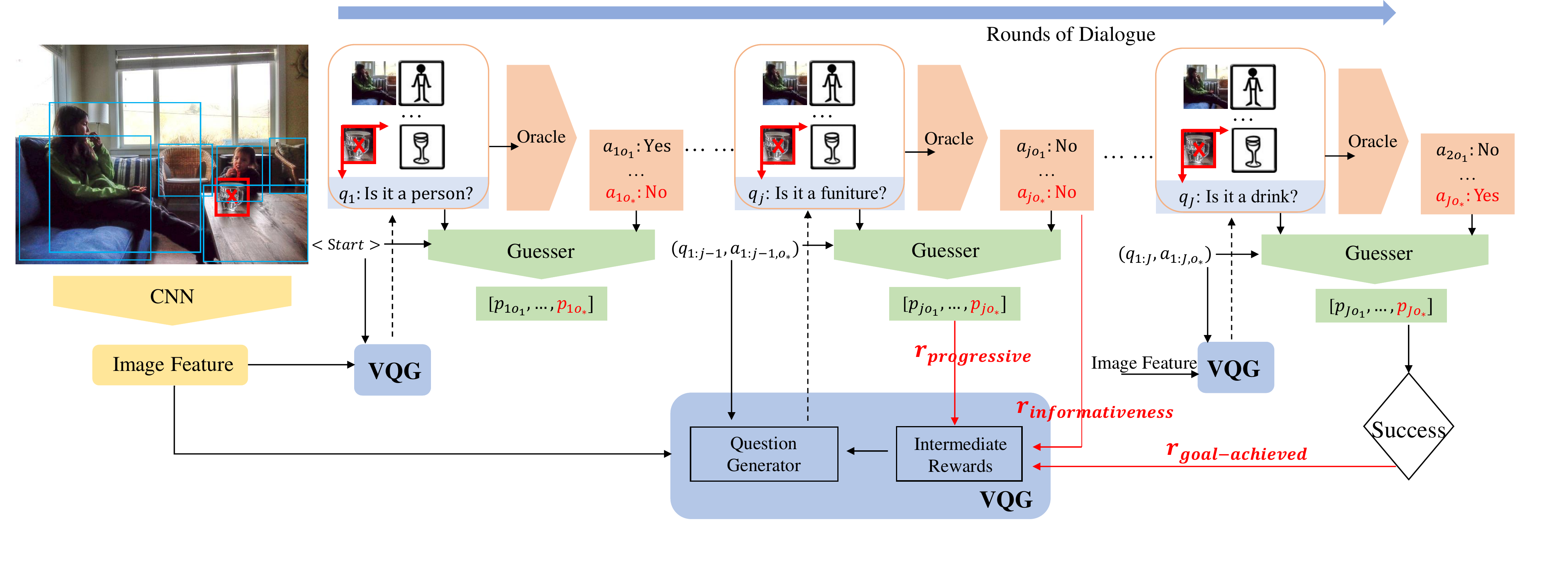}
	\vspace{-24pt}
	\caption{The framework of proposed VQG agent plays in the whole game environment. A target object $o^*$ is assigned to the Oracle, but it is unknown to VQG and Guesser. Then VQG generates a series of questions, which are answered by Oracle. During training, we let Oracle answer the question based on all the objects at each round, and measure the \textit{informativeness} reward, and we also let Guesser generate probability distribution to measure the \textit{progressive} reward. Finally, we consider the number of rounds $J$ and set the \textit{goal-achieved} reward based on the status of success. These intermediate rewards are adopted for optimizing the VQG agent by the REINFORCE.
	}
	\vspace{-12pt}
	\label{framework}
\end{figure*}

This is achieved through fitting the goal-oriented VQG into a reinforcement learning (RL) paradigm and devising three different intermediate rewards, which are our main contributions in this paper, to explicitly optimize the question generation. The first \textit{goal-achieved} reward is designed to encourage the agent to achieve the final goal (pick out the object that the \texttt{Oracle} is `thinking') via asking multiple questions. However, different from only considering whether the goal is achieved, additional rewards are awarded if the agent can use fewer questions to achieve it. This is a reasonable setting because you do not need a robot that can finish a task but has to ask you hundreds of questions. The second reward we proposed is the \textit{progressive} reward, which is established to encourage questions that generated by the agent can progressively increase the probability of the right answer. This is an intermediate reward for the individual question, and the reward is decided by the change of the ground-truth answer probability. A negative reward will be given if the probability decreases. The last reward is the \textit{informativeness} reward, which is used to restrict the agent not to ask `useless' questions, for example, a question that leads to the identical answer for all the candidate objects (this question cannot eliminate any ambiguous). We show the whole framework in Fig. \ref{framework}.

We evaluate our model on the GuessWhat?! dataset \cite{guesswhat_game}, with the pre-trained standard \texttt{Oracle} and \texttt{Guesser}, we show that our novel \texttt{Questioner} model outperforms the baseline and state-of-the-art model by a large margin. We also evaluate each reward respectively, to measure the individual contribution. Qualitative results show that we can produce more informative questions.

\section{Related Works}

\paragraph{Visual Question Generation} Recently, the visual question generation problem has been brought to the computer vision community, aims at generating visual-related questions. Most of the works treat the VQG as a standalone problem and follow an image captioning style framework, \ie, translate an image to a sentence, in this case, a question. For example, in \cite{mora2016towards}, Mora \etal use a CNN-LSTM model to generate questions and answers directly from the image visual content. Zhang \etal \cite{zhang2016automatic} focus on generating questions of grounded images. They use Densecap \cite{johnson2016densecap} as region captioning generator to guide the question generation. In \cite{mostafazadeh2016generating}, Mostafazadeh \etal propose a dataset to generate natural questions about images, which are beyond the literal description of image content. Li \etal \cite{li2017visual} view the VQA and VQG as a dual learning process by jointly training them in an end-to-end framework. Although these works can generate meaningful questions that are related to the image, the motivation of asking these questions are rather weak because they are not related to any goals. Moreover, it is hard to conduct the quality measurement on this type of questions. Instead, in our work, we aim to develop an agent that can learn to ask realistic questions, which can contribute to achieving a specific goal. 


Goal-oriented Visual Dialogue generation has attracted many attentions at most recently. In \cite{das2017learning}, Das \etal introduce a reinforcement learning mechanism for visual dialogue generation. They establish two RL agents corresponding to question and answer generation respectively, to finally locate an unseen image from a set of images. The question agent predicts the feature representation of the image and the reward function is given by measuring how close the representation is compared to the true feature. However, we focus on encouraging the agent to generate questions that directed towards the final goal, and we adopt different kinds of intermediate rewards to achieve that in the question generation process. Moreover, the question generation agent in their model only asks questions based on the dialogue history, which does not involve visual information. In \cite{end_to_end_gw}, Florian \etal propose to employ reinforcement learning to solve question generation of the GuessWhat game by introducing the final status of success as the sole reward. We share the similar backbone idea, but there are several technical differences. One of the most significant differences is that the previous work only considers using whether achieving the final goal as the reward but we assign different intermediate rewards for each posed question to push VQG agent to ask short but informative questions to achieve the goal. The experimental results and analysis in Section \ref{exp} show that our model not only outperforms the state-of-art but also achieves higher intelligence, \ie, using as few questions as possible to finish the task.

\vspace{-10pt}
\paragraph{Question Generation in NLP} There is a long history of works on grammar question generation from text domain in natural language processing (NLP) \cite{brown2005automatic,davey1986effects,singer1982active,therrien2006effect}. In \cite{agarwal2011automatic,becker2012mind}, authors focus on automatically generating gap-fill questions, while crowdsourcing templates and manually built templates are used for question generation in \cite{labutov2015deep} and \cite{mazidi2014linguistic} respectively. These works focus on constructing formatted questions from the text corpus.

\vspace{-10pt}
\paragraph{Reinforcement Learning for V2L} Reinforcement learning \cite{kaelbling1996reinforcement,sutton1998reinforcement} has been adopted in several vision to language (V2L) problems, including image captioning \cite{liu2016optimization,ren2017deep,rennie2016self}, VQA \cite{andreas2016learning,hu2017learning,zhuknowledge}, and aforementioned visual dialogue system \cite{das2017learning,lu2017best} \etc. In \cite{ren2017deep}, Ren \etal use a policy network and a value network to collaboratively generate image captions, while different optimization methods for RL in image captioning are explored in \cite{liu2016optimization} and \cite{rennie2016self}, called SPIDEr and self-critical sequence training. Zhu \etal \cite{zhuknowledge} introduce knowledge source into the iterative VQA and employ RL to learn the query policy. In \cite{andreas2016learning}, authors use RL to learn the parameters of QA model for both images and structured knowledge bases. 
These works solve V2L related problems by employing RL as an optimization method, while we focus on using RL with carefully designed intermediate rewards to train the VQG agent for goal-oriented tasks.
\section{Goal-Oriented VQG}

We ground our goal-oriented VQG problem on a \textit{Guess What} game, specifically, on the GuessWhat?! dataset~\cite{guesswhat_game}. GuessWhat?! is a three-role interactive game, where all roles observe the same image of a rich visual scene that contains multiple objects. We view this game as three parts: \texttt{Oracle}, \texttt{Questioner} and \texttt{Guesser}. In each game, a random object in the scene is assigned to the \texttt{Oracle}, where this process is hidden to the \texttt{Questioner}. Then the \texttt{Questioner} can ask a series of yes/no questions to locate this object. The list of objects is also hidden to the \texttt{Questioner} during the question-answer rounds. Once the \texttt{Questioner} has gathered enough information, the \texttt{Guesser} can start to guess. The game is considered as successful if the \texttt{Guesser} selects the right object.

The \texttt{Questioner} part of the game is a goal-oriented VQG problem; each question is generated based on the visual information of the image and the previous rounds of question-answer pairs. The goal of VQG is to successfully finish the game, in this case, to locate the right object. In this paper, we fit the goal-oriented VQG into a reinforcement learning paradigm and propose three different intermediate rewards, namely the \textit{goal-achieved} reward, \textit{progressive} reward, and \textit{informativeness} reward, to explicitly optimize the question generation. The \textit{goal-achieved} reward is established to lead the dialogue to achieve the final goal, the \textit{progressive} reward is used to push the intermediate generation process towards the optimal direction, while the \textit{informativeness} reward is used to ensure the quality of generated questions. To better express the generation process, we first introduce the notations of GuessWhat?! game, which is used throughout the rest of sections.

Each game is defined as a tuple $(I, D, O, o^*)$, where $I$ is the observed image, $D$ is the dialogue with $J$ rounds of question-answer pairs $(q_j,a_j)^{J}_{j=1}$, $O=(o_n)^{N}_{n=1}$ is the list of $N$ objects in the image $I$, where $o^*$ is the target object. Each question $q_j=(w_m^j)_{m=1}^{M_j}$ is a sequence of $M_j$ tokens, which are sampled from the pre-defined vocabulary $V$. The $V$ is composed of word tokens, a question stop token $<$?$>$ and a dialogue stop token $<$End$>$. The answer $a_j \in \{<$Yes$>$,$<$No$>$,$<$NA$>\}$ is set to be yes, no or not applicable. For each object $o$, it has an object category $c_o \in \{1\dots C\}$ and a segment mask.

\subsection{Learning Environment}

We build the learning environment to generate visual dialogues based on the GuessWhat?! dataset. Since we focus on the goal-oriented VQG, for a fair comparison, the \texttt{Oracle} and \texttt{Guesser} are produced by referring to the original baseline models in GuessWhat?! \cite{guesswhat_game}. We also introduce the VQG supervised learning model, which is referred as the baseline for the rest of the paper.

\vspace{-10pt}
\paragraph{Oracle}

The \texttt{Oracle} requires generating answers for all kinds of questions about any objects within the image scene. We build the neural network architecture for \texttt{Oracle} by referring to \cite{guesswhat_game}. The bounding box (obtained from the segment mask) of the object $o$ are encoded into an eight dimensional vector to represent the spatial feature, where $o_{spa}=[x_{min},y_{min},x_{max},y_{max},x_{center},y_{center},w_{box},h_{box}]$ indicates the box coordinates, width and height. The category $c_{o}$ is embedded using a learned look-up table, while the current question is encoded by an LSTM \cite{gers1999learning}. All three features are concatenated into a single vector and fed into a one hidden layer MLP followed by a softmax layer to produce the answer probability $p(a|o_{spa},c_o,q)$. 

\vspace{-10pt}
\paragraph{Guesser}

Given an image $I$ and a series of question-answer pairs, the \texttt{Guesser} requires predicting right object $o^*$ from a list of objects. By referring to \cite{guesswhat_game}, we consider the generated dialogue as one flat sequence of tokens and encode it with an LSTM. The last hidden state is extracted as the feature to represent the dialogue. We also embed all the objects' spatial features and categories by an MLP. We perform a dot-product between dialogue and object features with a softmax operation to produce the final prediction.

\vspace{-10pt}
\paragraph{VQG Baseline}

Given an image $I$ and a history of the question-answer pairs $(q,a)_{1:j-1}$, the VQG requires generating a new question $q_j$. We build the VQG baseline based on an RNN generator. The RNN recurrently produces a series of state vectors $s^j_{1:m}$ by transitioning from the previous state $s^j_{m-1}$ and the current input token $w_m^j$. We use an LSTM as the transition function $f$, that is, $s_m^j=f(s_{m-1}^j,w_m^j)$. In our case, the state vector $s$ is conditioned on the whole image and all the previous question-answer tokens. We add a softmax operation to produce the probability distribution over the vocabulary $V$, where $p(w_m^j|I,(q,a)_{1:j-1},w_{1:m-1}^j)$. This baseline is conducted by employing the supervised training. We train the VQG by minimizing the following negative log loss function:
\begin{equation}
\begin{aligned}
L&=-\log p(q_{1:J}|I,a_{1:J})\\
&=-\sum_{j=1}^{J}\sum_{m=1}^{M}\log p(w_m^j|I,w_{1:m-1}^j,(q,a)_{1:j-1})
\end{aligned}
\end{equation}
During the test stage, the question can be sampled from the model by starting from state $s_1^j$; a new token $w_m^j$ is sampled from the probability distribution, then embedded and fed back to the LSTM. We repeat this operation until the end of question token is encountered. 

\subsection{Reinforcement Learning of VQG}
We use our established \texttt{Oracle}, \texttt{Guesser} and VQG baseline model to simulate a complete GuessWhat?! game. Given an image $I$, an initial question $q_1$ is generated by sampling from the VQG baseline until the stop question token is encountered. Then the \texttt{Oracle} receives the question $q_1$ along with the assigned object category $o^*$ and its spatial information $o^*_{spa}$, and output the answer $a_1$, the question-answer pair $(q_1,a_1)$ is appended to the dialogue history. We repeat this loop until the end of dialogue token is sampled, or the number of questions reaches the maximum. Finally, the \texttt{Guesser} takes the whole dialogue $D$ and the object list $O$ as inputs to predict the object. We consider the goal reached if $o^*$ is selected. Otherwise, it failed.

To more efficiently optimize the VQG towards the final goal and generate informative questions, we adopt three intermediate rewards (which will be introduced in the following sections) into the RL framework.  

\vspace{-4pt}
\subsubsection{State, Action \& Policy}
\vspace{-2pt}
We view the VQG as a Markov Decision Process (MDP), the VQG is noted as the agent. For the dialogue generated based on the image $I$ at time step $t$, the state of agent is defined as the image visual content with the history of question-answer pairs and the tokens of current question generated so far: $S_t=(I,(q,a)_{1:j-1},(w_1^j,\dots ,w_m^j))$, where $t=\sum_{k=1}^{k=j-1}M_k+m$. The action $A_t$ of agent is to select the next output token $w^j_{m+1}$ from the vocabulary $V$. Depends on the actions that agent takes, the transition between two states falls into one of the following cases:

1) $w^j_{m+1}=<$?$>$: The current question is finished, the \texttt{Oracle} from the environment will answer $a_j$, which is appended to the dialogue history. The next state $S_{t+1}=(I,(q,a)_{1:j})$.

2) $w^j_{m+1}=<$End$>$: The dialogue is finished, the \texttt{Guesser} from the environment will select the object from the list $O$.

3) Otherwise, the new generated token $w^j_{m+1}$ keeps appending to the current question $q_j$, the next state $S_{t+1}=(I,(q,a)_{1:j-1},(w^j_1,\dots ,w^j_m,w^j_{m+1}))$.

The maximum length of question $q_j$ is $M_{max}$, and the maximum rounds of the dialogue is $J_{max}$. Therefore, the number of time steps $T$ of any dialogue are $T\leq M_{max}*J_{max}$. We model the VQG under the stochastic policy $\pi_\theta(A|S)$, where $\theta$ represents the parameters of the deep neural network we used in the VQG baseline that produces the probability distributions for each state. The goal of the policy learning is to estimate the parameter $\theta$.

After we set up the components of MDP, the most significant aspect of the RL is to define the appropriate reward function for each state-action pair $(S_t,A_t)$. As we emphasized before, the goal-oriented VQG aims to generate the questions that lead to achieving the final goal. Therefore, we build three kinds of intermediate rewards to push the VQG agent to be optimized towards the optimal direction. The whole framework is shown in Fig. \ref{framework}.

\vspace{-5pt}
\subsubsection{Goal-Achieved Reward}
\vspace{-4pt}
One basic rule of the appropriate reward function is that it cannot conflict with the final optimal policy \cite{ng1999policy}. The primary purpose of the VQG agent is to gather enough information as soon as possible to help \texttt{Guesser} to locate the object. Therefore, we define the first reward to reflect whether the final goal is achieved. Moreover, we take the number of rounds into consideration to accelerate the questioning part and let the reward nonzero when the game is successful.


Given the state $S_t$, where the $<$End$>$ token is sampled or the maximum number of rounds $J_{max}$ is reached, the reward of the state-action pair is defined as:
\begin{equation}
r_g(S_t,A_t)=\left\{
\begin{aligned}
&1\text{+}\lambda\cdot J_{max}/J,\text{ If \texttt{Guesser}}(S_t)=o^*\\
&0, \qquad \text{Otherwise}
\end{aligned}
\right.
\end{equation}
We set the reward as one plus the weighted maximum number of rounds $J_{max}$ against the actual rounds $J$ of the current dialogue if the dialogue is successful, and zero otherwise. This is based on that we want the final goal to motivate the VQG to generate useful questions. Moreover, the intermediate process is considered into the reward function as the rounds of the question-answer pairs $J$, which guarantees the efficiency of the generation process; the fewer questions are generated, the more reward VQG agent can get at the end of the game (if and only if the game succeed). This is a quite useful setting in the realistic because we do want to use fewer orders to guide the robot to finish more tasks. $\lambda$ is a weight to balance between the contribution of the successful reward and the dialogue round reward.

%
%
%
%

\vspace{-5pt}
\subsubsection{Progressive Reward}
\vspace{-4pt}
Based on the intuition and the observation of the human interactive dialogues, we find that the questions of a successful game, are ones that progressively achieve the final goal, \ie, as long as the questions being asked and answered, the confidence of referring to the target object becomes higher and higher. Therefore, at each round, we define an intermediate reward for state-action pair as the improvement of target probability that \texttt{Guesser} outputs. More specific, we interact with the \texttt{Guesser} at each round to obtain the probability of predicting target object. If the probability increases, it means that the generated question $q_j$ is a positive question that leads the dialogue towards the right direction. 

We set an intermediate reward called \textit{progressive} reward to encourage VQG agent progressively generate these positive questions. At each round $j$, we record the probability $p_j(o^*|I,(q,a)_{1:j})$ returned by Guesser, and compare it with the last round $j-1$. The difference between two probabilities is used as the intermediate reward. That is:
\begin{equation}
r_p(S_t,A_t)=p_j(o^*|I,(q,a)_{1:j})-p_{j-1}(o^*|I,(q,a)_{1:j-1})\\
\end{equation}
In this way, the question is considered high-quality and has a positive reward, if it leads to a higher probability to guess the right object. Otherwise, the reward is negative.

\subsubsection{Informativeness Reward}

When we human ask questions (especially in a guess what game), we expect an answer that can help us to eliminate the confusion and distinguish the candidate objects. Hence, imagine that if a posed question that leads to the same answer for all the candidate object, this question will be useless. For example, all the candidate objects are `red' and if we posed a question that `Is it red?', we will get the answer `Yes.' However, this question-answer pair cannot help us to identify the target. We want to avoid this kind of questions because they are non-informative. In this case, we need to evaluate the question based on the answer from the \texttt{Oracle}.  

Given generated question $q_j$, we interact with the \texttt{Oracle} to answer the question. Since the \texttt{Oracle} takes the image $I$, the current question $q_j$, and the target object $o^*$ as inputs, and outputs the answer $a_j$, we let the \texttt{Oracle} answer question $q_j$ for all objects in the image. If the answers are different from each other, we consider $q_j$ is useful for locating the right object. Otherwise, it does not contribute to the final goal. Therefore, we set the reward positive, which we called \textit{informativeness} reward, for these useful questions.

Formally, during each round, the \texttt{Oracle} receives the image $I$, the current question $q_j$ and the list of objects $O$, and then outputs the answer set $a_{jO}=\{a_{jo_1},\dots,a_{jo_{N}}\}$, where each element corresponds to each object. Then the \textit{informativeness} reward is defined as:
\vspace{-6pt}
\begin{equation}
r_i(S_t,A_t)=\left\{
\begin{aligned}
&\eta, \quad \text{If all $a_{jo_{n}}$ are not identical}\\
&0, \qquad \text{Otherwise}
\end{aligned}
\right.
\end{equation}
By giving a positive reward to the state-action pair, we improve the quality of the dialogue by encouraging agent to generate more informative questions.

\vspace{-5pt}
\subsubsection{Training with Policy Gradient}

Now we have three different kinds of rewards that take the intermediate process into consideration, for each state-action pair $(S_t,A_t)$, we add three rewards together as the final reward function:
\vspace{-6pt}
\begin{equation}
r(S_t,A_t)=r_g(S_t,A_t)+r_p(S_t,A_t)+r_i(S_t,A_t)
\end{equation}

Considering the large action space in the game setting, we adopt the policy gradient method \cite{sutton2000policy} to train the VQG agent with proposed intermediate rewards. The goal of policy gradient is to update policy parameters with respect to the expected return by gradient descent. Since we are in the episodic environment, given policy $\pi_\theta$, which is the generative network of the VQG agent, in this case, the policy objective function takes the form:
\vspace{-6pt}
\begin{equation}
J(\theta)=E_{\pi_\theta}[\sum_{t=1}^{T}r(S_t,A_t)]
\end{equation}
The parameters $\theta$ then can be optimized by following the gradient update rule. In REINFORCE algorithm \cite{kaelbling1996reinforcement}, the gradient of $J(\theta)$ can be estimated from a batch of episodes $\tau$ that are sampled from the policy $\pi_\theta$:
\vspace{-6pt}
\begin{equation}
\triangledown J(\theta)\approx \bigg\langle \sum_{t=1}^{T}\sum_{A_t\in V}\triangledown_{\theta} \log\pi_\theta (S_t,A_t)(Q^{\pi_\theta}(S_t,A_t)-b_{\varphi})\bigg\rangle_{\tau}
\end{equation}
where $Q^{\pi_\theta}(S_t,A_t)$ is the state-action value function that returns the expectation of cumulative reward at $(S_t,A_t)$:
\vspace{-6pt}
\begin{equation}
Q^{\pi_\theta}(S_t,A_t)=E_{\pi_\theta}[\sum_{t'=t}^{T}r(S_{t'},A_{t'})]
\end{equation}
by substituting the notations with VQG agent, we have the following policy gradient:
\vspace{-6pt}
\begin{equation}
\label{eq1}
\begin{aligned}
\triangledown J(\theta)\approx \bigg\langle\sum_{j=1}^{J}\sum_{m=1}^{M_j}\triangledown_{\theta} \log\pi_\theta (w_m^j|I,(q,a)_{1:j-1},w_{1:m-1}^j)\\
(Q^{\pi_\theta}(I,(q,a)_{1:j-1},w_{1:m-1}^j,w_m^j)-b_{\varphi})\bigg\rangle_{\tau}
\end{aligned}
\end{equation}
$b_{\varphi}$ is a baseline function to help reduce the gradient variance, which can be chosen arbitrarily. We use a one-layer MLP that takes state $S_t$ as input in VQG agent and outputs the expected reward. The baseline $b_{\varphi}$ is trained with mean squared error as:
\vspace{-7pt}
\begin{equation}
\label{eq2}
\min_{\varphi} L(\varphi)=\bigg\langle[b_{\varphi}(S_t)-\sum_{t'=t}^{T}r(S_{t'},A_{t'})]^2\bigg\rangle_{\tau}
\end{equation}
The whole training procedure is shown in Alg.\ref{input generation}.

\begin{algorithm}[t] 
	\caption{Training procedure of the VQG agent.}
	\label{input generation} 
	\small
	\begin{algorithmic}[1]
		\Require \texttt{Oracle}$(Ora)$, \texttt{Guesser}$(Gus)$, $VQG$, batch size $H$
		\For{Each update}
		\State \# Generate episodes $\tau$
		\For{$h=1$ to $H$}
		\State select image $I_h$ and one target object $o_h^*\in O_h$
		\State \# Generate question-answer pairs $(q,a)_{1:j}^h$
		\For {$j=1$ to $J_{max}$}
		\State $q_j^h=VQG(I_h,(q,a)^h_{1:j-1})$
		\State \# $N$ is the number of total objects
		\For {$n=1$ to $N$}
		\State $a^h_{jo_{hn}}=Ora(I_h,q_j^h,o_{hn})$
		\EndFor
		\If {all $a^h_{jo_{hn}}$ are not identical}
		\State $r_i(S_t,A_t)=\eta$
		\Else { $r_i(S_t,A_t)=0$}
		\EndIf
		\State $r(S_t,A_t)=r_i(S_t,A_t)$
		\State $p_j(o^*_{h}|\cdot)=Gus(I_h,(q,a)^h_{1:j},O_h)$
		\If {$j>1$}
		\State $r_p(S_t,A_t)=p_j(o^*_{h}|\cdot)-p_{j-1}(o^*_{h}|\cdot)$
		\State $r(S_t,A_t)=r(S_t,A_t)+r_p(S_t,A_t)$
		\EndIf
		\If {$<$End$>\in q_j^h$}
		\State break;
		\EndIf
		\EndFor
		\State $p(o^h|\cdot)=Gus(I_h,(q,a)^h_{1:j},O_h)$
		\If {$argmax_{o_h}p(o^h|\cdot)=o_h^*$}
		\State $r_g(S_t,A_t)=1+\lambda \cdot J_{max}/j$
		\Else { $r_g(S_t,A_t)=0$}
		\EndIf
		\State $r(S_t,A_t)=r(S_t,A_t)+r_g(S_t,A_t)$
		\EndFor
		\State Define $\tau=(I_h,(q,a)^h_{1:j_h},r_h)_{1:H}$
		\State Evaluate $\triangledown J(\theta)$ as Eq. \ref{eq1} and update VQG agent
		\State Evaluate $\triangledown L(\varphi)$ as Eq. \ref{eq2} and update $b_{\varphi}$ baseline
		\EndFor
	\end{algorithmic}
\end{algorithm}
\vspace{-5.7pt}
\section{Experiment}
\label{exp}
In this section, we present our VQG results and conduct comprehensive ablation analysis about each intermediate reward. As mentioned above, the proposed method is evaluated on the GuessWhat?! game dataset \cite{guesswhat_game} with pre-trained standard \texttt{Oracle} and \texttt{Guesser}. By comparing with the baseline and the state-of-the-art model, we show that proposed model can efficiently generate informative questions, which serve the final goal.

\subsection{Dataset \& Evaluation Metric}
The GuessWhat?! Dataset \cite{guesswhat_game} is composed of 155,281 dialogues grounded on the 66,537 images with 134,074 unique objects. There are 821,955 question-answer pairs in the dialogues with vocabulary size 4,900. We use the standard split of training, validation and test in \cite{guesswhat_game,end_to_end_gw}.

Following \cite{end_to_end_gw}, we report the accuracies of the games as the evaluation metric. Given a $J$-round dialogue, if the target object $o^*$ is located by \texttt{Guesser}, the game is noted as successful, which indicates that the VQG agent has generated the qualified questions to serve the final goal. There are two kinds of test runs on the training set and test set respectively, named as \textbf{NewObject} and \textbf{NewImage}. NewObject is randomly sampling target objects from the training images (but we restrict only to use new objects that are not seen before), while NewImage is sampling objects from the test images (unseen). We report three inference methods namely sampling, greedy and beam-search (beam size is 5) for these two test runs.

\subsection{Implementation Details}
The standard \texttt{Oracle}, \texttt{Guesser} and VQG baseline are reproduced by referring to \cite{end_to_end_gw}. The error of trained \texttt{Oracle}, \texttt{Guesser} on test set are 21.1\% and 35.8\% respectively. The VQG baseline is referred as Baseline in Tab.\ref{NewObject} and \ref{NewImage} \footnote{These results are reported on https://github.com/GuessWhatGame by original authors.}.

We initialize the training environment with the standard \texttt{Oracle}, \texttt{Guesser} and VQG baseline, then start to train the VQG agent with proposed reward functions. We train our models for 100 epochs with stochastic gradient descent (SGD) \cite{bottou2010large}. The learning rate and batch size are 0.001 and 64, respectively. The baseline function $b_{\varphi}$ is trained with SGD at the same time. During each epoch, each training image is sampled once, and one of the objects inside it is randomly assigned as the target. We set the maximum round $J_{max}=5$ and maximum length of question $M_{max}=12$. The weight of the dialog round reward is set to $\lambda=0.1$. The progressive reward is set as $\eta=0.1$\footnote{We use a grid search to select the hyper-parameters $\lambda$ and $\eta$, we find 0.1 produces the best results.}.

\subsection{Results \& Ablation Analysis}
In this section, we give the overall analysis on proposed intermediate reward functions. To better show the effectiveness of each reward, we conduct comprehensive ablation studies. Moreover, we also carry out a human interpretability study to evaluate whether human subjects can understand the generated questions and how well the human can use these question-answer pairs to achieve the final goal. We note VQG agent trained with \textit{goal-achieved} reward as VQG-$r_g$, trained with \textit{goal-achieved} and \textit{progressive} rewards as VQG-$r_g$+$r_p$, trained with \textit{goal-achieved} and \textit{informativeness} rewards as VQG-$r_g$+$r_i$. The final agent trained with all three rewards is noted as VQG-$r_g$+$r_p$+$r_i$.

\vspace{-10pt}
\paragraph{Overall Analysis}

Tab. \ref{NewObject} and \ref{NewImage} show the comparisons between VQG agent optimized by proposed intermediate rewards and the state-of-the-art model proposed in \cite{end_to_end_gw} noted as Sole-$r$, which uses indicator of whether reaching the final goal as the sole reward function. As we can see, with proposed intermediate rewards and their combinations, our VQG agents outperform both compared models on all evaluation metrics. More specifically, our final VQG-$r_g$+$r_p$+$r_i$ agent surpasses the Sole-$r$ 4.7\%, 3.3\% and 3.7\% accuracy on NewObject sampling, greedy and beam-search respectively, while obtains 3.3\%, 2.3\% and 2.4\% higher accuracy on NewImage sampling, greedy and beam-search respectively. Moreover, all of our agents outperform the supervised baseline by a significant margin.

\begin{table}[t]
	\centering
	\vspace{-5pt}
	\caption{Results on training images (NewObject).}
	\vspace{-5pt}
	\label{NewObject}
	\begin{tabular}{c|ccc}
		\hline
		NewObject   & Sampling & Greedy & Beam-Search   \\ \hline
		Baseline \cite{guesswhat_game}   & 41.6     & 43.5   & 47.1                         \\ \
		Sole-$r$ \cite{end_to_end_gw}         & 58.5     & 60.3   & 60.2                         \\ \hline
		VQG-$r_g$     &60.6          &61.7        &61.4                              \\ 
		VQG-$r_g$+$r_p$   &62.1          &62.9        &63.1                              \\ 
		VQG-$r_g$+$r_i$   &61.3          &62.4        &62.7                              \\ 
		VQG-$r_g$+$r_p$+$r_i$ &\textbf{63.2}          &\textbf{63.6}        &\textbf{63.9}                              \\ \hline
	\end{tabular}
\end{table}

\begin{table}[t]
	\centering
	\vspace{-5pt}
	\caption{Results on test images (NewImage).}
	\vspace{-5pt}
	\label{NewImage}
	\begin{tabular}{c|ccc}
		\hline
		NewImage   & Sampling & Greedy & Beam-Search  \\ \hline
		Baseline \cite{guesswhat_game}   & 39.2     & 40.8   & 44.6                         \\ 
		Sole-$r$ \cite{end_to_end_gw}         & 56.5     & 58.4   & 58.4                         \\ \hline
		VQG-$r_g$     &58.2          &59.3        &59.4                              \\ 
		VQG-$r_g$+$r_p$   &59.3          &60.6        &60.5                              \\ 
		VQG-$r_g$+$r_i$   &58.5          &59.7        &60.1                             \\ 
		VQG-$r_g$+$r_p$+$r_i$ &\textbf{59.8}          &\textbf{60.7}        &\textbf{60.8}                              \\ \hline
	\end{tabular}
\vspace{-10pt}
\end{table}

\begin{figure*}[t!]
	\centering
	\includegraphics[width=1\linewidth]{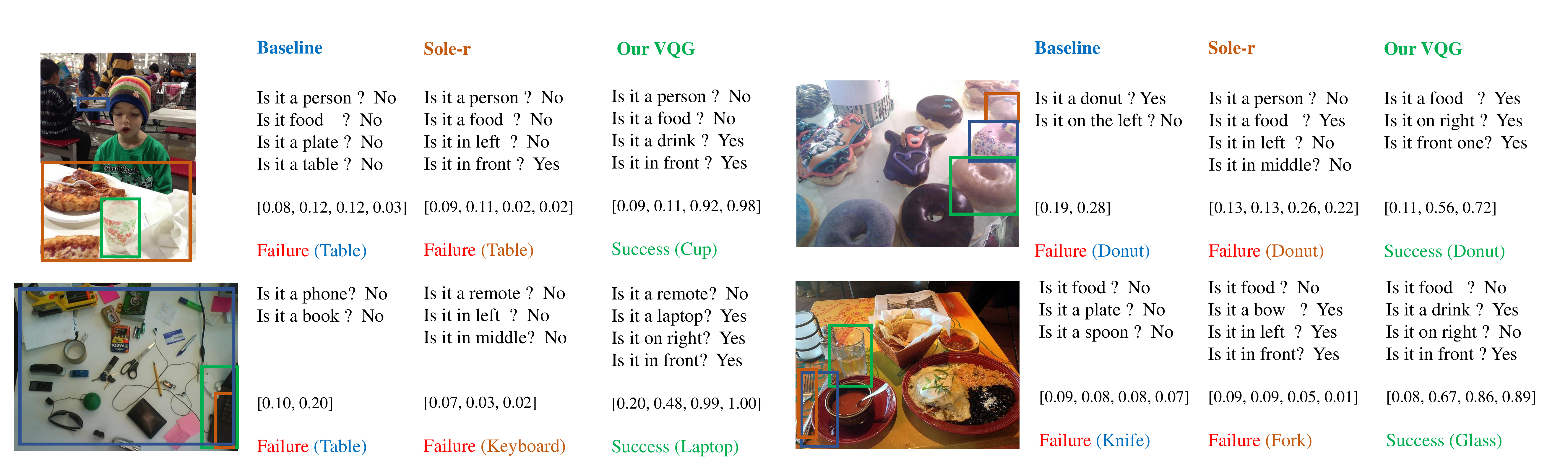}
	\caption{Some qualitative results of our VQG agent (green), and the comparisons with baseline (blue) and Sole-$r$ model (brown). The elements in the middle array indicate the probabilities of successfully locating the target object after each round. Better viewed in color.
	}
	\label{examples}
	\vspace{-12pt}
\end{figure*}

To fully show the effectiveness of our proposed intermediate rewards, we train three VQG agents using $r_g$, $r_g$+$r_p$, and $r_g$+$r_i$ rewards respectively, and conduct ablation analysis. As we can see, the VQG-$r_g$ already outperforms both baseline and the state-of-the-art model, which means that controlling dialogue round can push the agent to ask more wise questions. With the combination of $r_p$ and $r_i$ reward respectively, the performance of VQG agent further improved. We find that the improvement gained from $r_p$ reward is higher than $r_i$ reward, which suggests that the intermediate \textit{progressive} reward contributes more in our experiment. Our final agent combines all rewards and achieves the best results. Fig. \ref{examples} shows some qualitative results. More results can be found in the supplementary material, including some fail cases.

\begin{figure}[t!]
	\centering
	\includegraphics[width=0.9\linewidth]{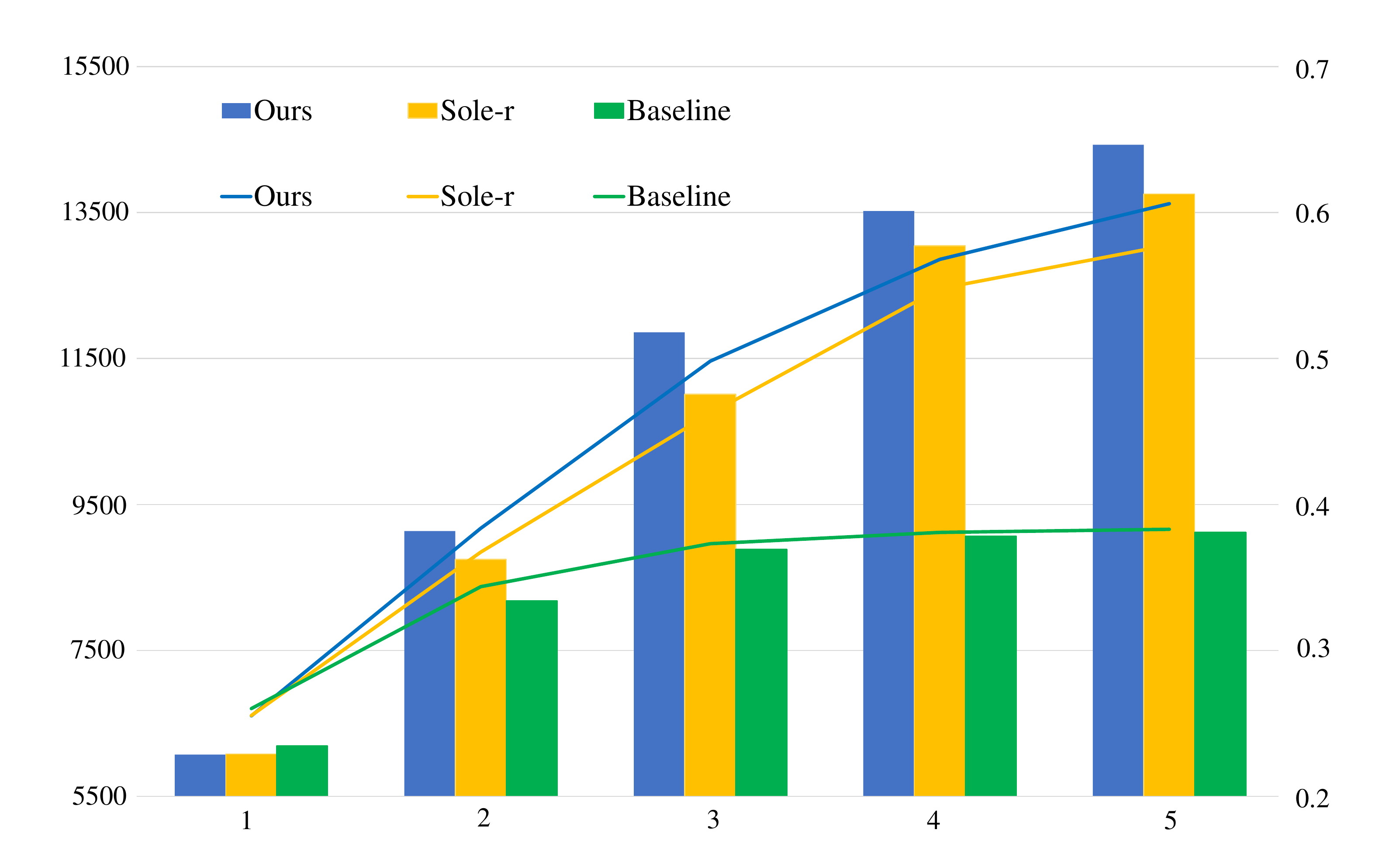}
	\caption{The comparisons of success ratio between our agent and Sole-$r$, as well the baseline model, at different dialogue round. The left y-axis indicates the number of successful dialogues, which corresponds to the bar chart. The right y-axis indicates the success ratio, which corresponds to the line chart. Better viewed in color.
	}
	\label{dialogue_len}
	\vspace{-12pt}
\end{figure}

\vspace{-12pt}
\paragraph{Dialogue Round}
We conduct an experiment to investigate the relationship between the dialogue round and the game success ratio. More specifically, we let \texttt{Guesser} to select the object at each round and calculate the success ratio at the given round, the comparisons of different models are shown in Fig. \ref{dialogue_len}. As we can see, our agent can achieve the goal at fewer rounds compared to the other models, especially at the round three.

\vspace{-12pt}
\paragraph{Progressive Trend}
To prove our VQG agent can learn a progressive trend on generated questions, we count the percentage of the successful game that has a progressive (ascending) trend on the target object, by observing the probability distributions generated by \texttt{Guesser} at each round. Our agent achieves 60.7\%, while baseline and Sole-$r$ are 50.8\% and 57.3\% respectively, which indicates that our agent is better at generating questions in a progressive trend considering we introduce the progressive reward, $r_p$. Some qualitative results of the `progressive trend' are shown in the Fig. \ref{examples}, \ie, the probability of the right answer is progressively increasing.

\vspace{-12pt}
\paragraph{Question Informativeness}
We also investigate the informativeness of the questions generated by different models. We let \texttt{Oracle} answer questions for all the objects at each round, and count the percentage of high-quality questions in the successful game. We define that a high-quality question is a one does not lead to the same answer for all the candidate objects. The experimental results show that our VQG agent has 87.7\% high-quality questions, which is higher than the baseline (84.7\%) and Sole-$r$ (86.3\%). This confirms the contribution of the $r_i$ reward.

\subsection{Human Study}
We conduct a human study to see how well human can guess the target object based on the questions generated by these models. We show human subjects 50 images with generated question-answer pairs from baseline, Sole-$r$, and our final VQG agent, and let them guess the objects, \ie, replacing the AI \texttt{guesser} to a real human. We ask three human subjects to play on the same split, and the game is recognized as successful if at least two of them give the right answer. Based on our experiments, averagely, subjects achieve the highest accuracy 76\% based on our agent, which achieves 52\% and 70\% accuracies on the baseline and Sole-$r$ questions respectively. These results indicate that our agent can generate higher qualitative questions that can benefit the human to achieve the final goal.

\vspace{-6pt}
\section{Conclusion}
The ability to devise concise questions that lead to two parties to a dialog satisfying a shared goal as effectively as possible has important practical applications and theoretical implications. By introducing suitably crafted intermediate rewards into a deep reinforcement learning framework, we have shown that it is possible to achieve this result, at least for a particular class of goal.

The method we have devised not only achieves the Guess What goal reliably and succinctly but also outperforms the state-of-art. However, since the \texttt{Oracle} and \texttt{Guesser} are fixed, they are inaccurate to a certain extent. Consider the main objective of this paper is to show the effectiveness of our proposed intermediate rewards on the VQG problem, we leave it as the further work to train the three components jointly, with a reinforcement learning framework.

%

{\small
	\bibliographystyle{ieee}
	\bibliography{egbib}

\begin{thebibliography}{10}\itemsep=-1pt

\bibitem{agarwal2011automatic}
M.~Agarwal and P.~Mannem.
\newblock Automatic gap-fill question generation from text books.
\newblock In {\em Proc. Workshop Inno. Use of NLP for Buil. Educ. Appl.}, pages
  56--64. Association for Computational Linguistics, 2011.

\bibitem{andreas2016learning}
J.~Andreas, M.~Rohrbach, T.~Darrell, and D.~Klein.
\newblock Learning to compose neural networks for question answering.
\newblock {\em arXiv preprint arXiv:1601.01705}, 2016.

\bibitem{antol2015vqa}
S.~Antol, A.~Agrawal, J.~Lu, M.~Mitchell, D.~Batra, C.~Lawrence~Zitnick, and
  D.~Parikh.
\newblock Vqa: Visual question answering.
\newblock In {\em Proceedings of the IEEE International Conference on Computer
  Vision}, pages 2425--2433, 2015.

\bibitem{becker2012mind}
L.~Becker, S.~Basu, and L.~Vanderwende.
\newblock Mind the gap: learning to choose gaps for question generation.
\newblock In {\em Proc. Conf. North Amer. Chap. Asso. for Comp. Ling.: Human
  Lang. Tech.}, pages 742--751. Association for Computational Linguistics,
  2012.

\bibitem{bottou2010large}
L.~Bottou.
\newblock Large-scale machine learning with stochastic gradient descent.
\newblock In {\em Proceedings of COMPSTAT'2010}, pages 177--186. Springer,
  2010.

\bibitem{brown2005automatic}
J.~C. Brown, G.~A. Frishkoff, and M.~Eskenazi.
\newblock Automatic question generation for vocabulary assessment.
\newblock In {\em Proc. Conf. Human Lang. Tech. and Empi. Meth. in Natu. Lang.
  Process}, pages 819--826. Association for Computational Linguistics, 2005.

\bibitem{das2016visual}
A.~Das, S.~Kottur, K.~Gupta, A.~Singh, D.~Yadav, J.~M. Moura, D.~Parikh, and
  D.~Batra.
\newblock Visual dialog.
\newblock {\em arXiv preprint arXiv:1611.08669}, 2016.

\bibitem{das2017learning}
A.~Das, S.~Kottur, J.~M. Moura, S.~Lee, and D.~Batra.
\newblock Learning cooperative visual dialog agents with deep reinforcement
  learning.
\newblock {\em arXiv preprint arXiv:1703.06585}, 2017.

\bibitem{davey1986effects}
B.~Davey and S.~McBride.
\newblock Effects of question-generation training on reading comprehension.
\newblock {\em Journal of Educational Psychology}, 78(4):256, 1986.

\bibitem{guesswhat_game}
H.~de~Vries, F.~Strub, S.~Chandar, O.~Pietquin, H.~Larochelle, and A.~C.
  Courville.
\newblock Guesswhat?! visual object discovery through multi-modal dialogue.
\newblock In {\em {Proc. IEEE Conf. Comp. Vis. Patt. Recogn.}}, 2017.

\bibitem{gers1999learning}
F.~A. Gers, J.~Schmidhuber, and F.~Cummins.
\newblock Learning to forget: Continual prediction with lstm.
\newblock 1999.

\bibitem{hu2017learning}
R.~Hu, J.~Andreas, M.~Rohrbach, T.~Darrell, and K.~Saenko.
\newblock Learning to reason: End-to-end module networks for visual question
  answering.
\newblock {\em arXiv preprint arXiv:1704.05526}, 2017.

\bibitem{johnson2016densecap}
J.~Johnson, A.~Karpathy, and L.~Fei-Fei.
\newblock Densecap: Fully convolutional localization networks for dense
  captioning.
\newblock In {\em {Proc. IEEE Conf. Comp. Vis. Patt. Recogn.}}, pages
  4565--4574, 2016.

\bibitem{kaelbling1996reinforcement}
L.~P. Kaelbling, M.~L. Littman, and A.~W. Moore.
\newblock Reinforcement learning: A survey.
\newblock {\em J. Arti. Intell. Research}, 4:237--285, 1996.

\bibitem{labutov2015deep}
I.~Labutov, S.~Basu, and L.~Vanderwende.
\newblock Deep questions without deep understanding.
\newblock In {\em ACL (1)}, pages 889--898, 2015.

\bibitem{li2017visual}
Y.~Li, N.~Duan, B.~Zhou, X.~Chu, W.~Ouyang, and X.~Wang.
\newblock Visual question generation as dual task of visual question answering.
\newblock {\em arXiv preprint arXiv:1709.07192}, 2017.

\bibitem{liu2016optimization}
S.~Liu, Z.~Zhu, N.~Ye, S.~Guadarrama, and K.~Murphy.
\newblock Optimization of image description metrics using policy gradient
  methods.
\newblock {\em arXiv preprint arXiv:1612.00370}, 2016.

\bibitem{lu2017best}
J.~Lu, A.~Kannan, J.~Yang, D.~Parikh, and D.~Batra.
\newblock Best of both worlds: Transferring knowledge from discriminative
  learning to a generative visual dialog model.
\newblock {\em arXiv preprint arXiv:1706.01554}, 2017.

\bibitem{mazidi2014linguistic}
K.~Mazidi and R.~D. Nielsen.
\newblock Linguistic considerations in automatic question generation.
\newblock In {\em ACL (2)}, pages 321--326, 2014.

\bibitem{mora2016towards}
I.~M. Mora, S.~P. de~la Puente, and X.~Giro-i Nieto.
\newblock Towards automatic generation of question answer pairs from images,
  2016.

\bibitem{mostafazadeh2016generating}
N.~Mostafazadeh, I.~Misra, J.~Devlin, M.~Mitchell, X.~He, and L.~Vanderwende.
\newblock Generating natural questions about an image.
\newblock {\em arXiv preprint arXiv:1603.06059}, 2016.

\bibitem{ng1999policy}
A.~Y. Ng, D.~Harada, and S.~Russell.
\newblock Policy invariance under reward transformations: Theory and
  application to reward shaping.
\newblock In {\em {Proc. Int. Conf. Mach. Learn.}}, volume~99, pages 278--287,
  1999.

\bibitem{ren2017deep}
Z.~Ren, X.~Wang, N.~Zhang, X.~Lv, and L.-J. Li.
\newblock Deep reinforcement learning-based image captioning with embedding
  reward.
\newblock {\em {Proc. IEEE Conf. Comp. Vis. Patt. Recogn.}}, 2017.

\bibitem{rennie2016self}
S.~J. Rennie, E.~Marcheret, Y.~Mroueh, J.~Ross, and V.~Goel.
\newblock Self-critical sequence training for image captioning.
\newblock {\em arXiv preprint arXiv:1612.00563}, 2016.

\bibitem{singer1982active}
H.~Singer and D.~Donlan.
\newblock Active comprehension: Problem-solving schema with question generation
  for comprehension of complex short stories.
\newblock {\em Reading Research Quarterly}, pages 166--186, 1982.

\bibitem{end_to_end_gw}
F.~Strub, H.~de~Vries, J.~Mary, B.~Piot, A.~C. Courville, and O.~Pietquin.
\newblock End-to-end optimization of goal-driven and visually grounded dialogue
  systems.
\newblock In {\em {Proc. Int. Joint Conf. Artificial Intell.}}, 2017.

\bibitem{sutton1998reinforcement}
R.~S. Sutton and A.~G. Barto.
\newblock {\em Reinforcement learning: An introduction}, volume~1.
\newblock MIT press Cambridge, 1998.

\bibitem{sutton2000policy}
R.~S. Sutton, D.~A. McAllester, S.~P. Singh, and Y.~Mansour.
\newblock Policy gradient methods for reinforcement learning with function
  approximation.
\newblock In {\em Advances in neural information processing systems}, pages
  1057--1063, 2000.

\bibitem{therrien2006effect}
W.~J. Therrien, K.~Wickstrom, and K.~Jones.
\newblock Effect of a combined repeated reading and question generation
  intervention on reading achievement.
\newblock {\em Learning Disabilities Research \& Practice}, 21(2):89--97, 2006.

\bibitem{Wu_2016_CVPR}
Q.~Wu, P.~Wang, C.~Shen, A.~Dick, and A.~van~den Hengel.
\newblock Ask me anything: Free-form visual question answering based on
  knowledge from external sources.
\newblock In {\em {Proc. IEEE Conf. Comp. Vis. Patt. Recogn.}}, June 2016.

\bibitem{xu2016ask}
H.~Xu and K.~Saenko.
\newblock Ask, attend and answer: Exploring question-guided spatial attention
  for visual question answering.
\newblock In {\em {Proc. Eur. Conf. Comp. Vis.}}, pages 451--466. Springer,
  2016.

\bibitem{zhang2016automatic}
S.~Zhang, L.~Qu, S.~You, Z.~Yang, and J.~Zhang.
\newblock Automatic generation of grounded visual questions.
\newblock {\em arXiv preprint arXiv:1612.06530}, 2016.

\bibitem{zhuknowledge}
Y.~Zhu, J.~J. Lim, and L.~Fei-Fei.
\newblock Knowledge acquisition for visual question answering via iterative
  querying.
\newblock {\em {Proc. IEEE Conf. Comp. Vis. Patt. Recogn.}}, 2017.

\end{thebibliography}
}

\end{document}